\documentclass{article}
\usepackage{spconf,amsmath,graphicx}
\usepackage{multicol}
\usepackage{graphicx}
\usepackage{multirow}
\usepackage{amssymb,amsmath,bm}
\usepackage{enumitem}

\usepackage{array}

\ninept

\title{On Modular Training of Neural Acoustics-to-Word Model for LVCSR}
\name{Zhehuai Chen, Qi Liu, Hao Li and Kai Yu
\thanks{
The corresponding author is Kai Yu. This work has been supported by the National Key Research and Development Program of China under Grant No.2017YFB1002102. Experiments have been carried out on the PI supercomputer at Shanghai Jiao Tong University.
}
}
\address{
Key Lab. of Shanghai Education Commission for Intelligent Interaction and Cognitive Engineering \\
SpeechLab, Department of Computer Science and Engineering, Shanghai Jiao Tong University, China \\
}
\begin{document}
%
\maketitle
\begin{abstract}
End-to-end (E2E) automatic speech recognition (ASR) systems directly map acoustics to words using a unified model. Previous works mostly focus on E2E training a single model which integrates acoustic and language model into a whole. Although E2E training benefits from sequence modeling and simplified decoding pipelines, large amount of transcribed acoustic data is usually required, and traditional acoustic and language modelling techniques cannot be utilized. In this paper, a novel modular training framework of E2E ASR is proposed to separately train neural acoustic and language models during training stage, while still performing end-to-end inference in decoding stage. Here, an acoustics-to-phoneme model (A2P) and a phoneme-to-word model (P2W) are trained using acoustic data and text data respectively. A phone synchronous decoding (PSD) module is inserted between A2P and P2W to reduce sequence lengths without precision loss. Finally, modules are integrated into an acoustics-to-word model (A2W) and jointly optimized using acoustic data to retain the advantage of sequence modeling. Experiments on a 300-hour Switchboard task show significant improvement over the direct A2W model. The efficiency in both training and decoding also benefits from the proposed method.
%
%
\end{abstract}
\vspace{-0.5em}           
\section{Introduction}
\label{sec:intro}
\vspace{-0.5em}
Acoustic modeling with deep neural network (DNN) commonly operates in the
hybrid framework: DNNs, as discriminative
models, estimate the posterior probabilities of Hidden Markov Model (HMM) states, while  DNN and HMM are optimized separately.
Recent advances in more powerful neural networks enable stronger modeling effects in the context and history of sequence modeling~\cite{sak2014long,qian2016very}.
More labeled data further alleviates the sparseness and generalization problem in the modeling.
Thus in automatic speech recognition (ASR), it is promising to model the acoustics in sequence level without frame decomposition and separate optimization.
End-to-end (E2E) automatic speech recognition (ASR) systems directly map acoustics to words using a unified model, which integrates acoustic and language model into a whole. 
E2E usually  utilizes larger inference granularities, which simplifies the ASR pipeline and reduces the decoding complexity, e.g. removing the lexicon design, and inferring word or grapheme sequences directly. E2E can be divided into two branches as below:

Connectionist temporal classification (CTC)  is proposed to
model the posterior probability of the label sequence given the acoustic feature sequence.
The $\tt blank$ label unit is introduced to predict  the label sequence on unsegmented data 
 at any time step in the input sequence.
%
Researches have been conducted on different model granularities from tri-phone to word~\cite{sak2015fast,soltau2016neural}.
Recently, novel structures and respective criteria~\cite{graves2012sequence,sak2017recurrent,tang2017end} are proposed to improve the conditional independent assumption (CIA) in CTC.
Attention based encoder-decoder (S2S)~\cite{collobert2016wav2letter,chan2016end} is another end-to-end model without CIA. It  predicts the posterior probability of the label sequence given the acoustic feature sequence and previous inference labels.
An attention mechanism weights the hidden vectors of the feature sequence to use the most related hidden vectors  for  prediction.

Both E2E optimization and decoding are operated in these models.
Although E2E ASR benefits from sequence modeling, large amount of acoustic data is required  to  optimize the model jointly, 
e.g., a word level CTC is trained on 125K hours labeled data and outperforms hybrid models~\cite{soltau2016neural}. 
Meanwhile, the end-to-end decoding reduces the  weighted finite state transducer (WFST) based search complexity~\cite{zhc00-chen-tasl2017}, which is one of the key advantages.

In this paper, a modular training strategy is proposed  to utilize external sources to train each building block, while the end-to-end decoding is retained.
The ideas we apply to E2E systems are:
i) utilizing acoustic and text data in end-to-end ASR modeling by modular training strategy. A CTC acoustics-to-phoneme model (A2P)   is trained using acoustic data. Text data is utilized to train a phoneme-to-word model (P2W)  of  CTC or S2S.
ii) combining modules into an acoustics-to-word model (A2W)  by phone synchronous decoding (PSD)~\cite{zhc00-chen-tasl2017} and joint optimization. Namely, a PSD module is proposed on  CTC inference sequences to skip $\tt blank$-dominated frames, which reduces the sequence length without precision loss.
Finally, the A2P, PSD and P2W modules are stacked and jointly fine-tuned using acoustic data to retain the advantage of sequence modeling.
The advantages include: i) easier and faster model convergence due to modularization and initialization. ii) easy to utilize traditional AM and language model (LM) technologies  using  text  and acoustic data respectively. 

The rest of the paper is organized as follows. In Section~\ref{sec:e2e-review}, end-to-end training methods are briefly reviewed. In Section~\ref{sec:modu}, the modular training framework is proposed and compared with prior works in Section~\ref{sec:relatework}. In Section~\ref{sec:exp}, experiments are conducted on Switchboard corpus in the context of both CTC and S2S. Finally we present our conclusions in Section~\ref{sec:conclu}.

%
%
%
%
%

\vspace{-0.8em}  
\section{End-to-end Speech Recognition}
\label{sec:e2e-review}
\vspace{-0.5em}

\subsection{Connectionist Temporal Classification}
\label{sec:ctc-review}
\vspace{-0.5em}  
CTC~\cite{graves2006connectionist} directly calculates the posterior probability $P(\mathbf{l}|\mathbf{x})$ of the sequence $\mathbf{l}$ given the feature sequence $\mathbf{x}$. It is proposed to label unsegmented data by introduction of the $\tt blank$ label unit to  predict the label sequence at any time step in the input sequence.
\begin{equation}
\label{equ:ctc-model}
\begin{split}
P(\mathbf{l}|\mathbf{x})=\sum_{\pi\in\mathcal{B}^{-1}(\mathbf{l})}P(\pi|\mathbf{x}) =\sum_{\pi}\prod_{t=1}^{T} P(\pi_t|\mathbf{x})
\end{split}
\end{equation}
where $\mathcal{B}$ is a many-to-one mapping defined as below.
\begin{equation}
\label{equ:ctc-b}
\begin{split}
\mathcal{B}:   {L} \cup \{{\tt blank}\}  \mapsto {L}
\end{split}
\end{equation}
$\mathcal{B}$ determines the label sequence $\mathbf{l}$ and its corresponding set of CTC paths $\pi$. The mapping is by inserting an optional and self-loop $ \tt blank$ between each label unit $l$ in $\mathbf{l}$. $P(\pi_t|\mathbf{x})$ is estimated from the neural network taking the feature sequence $\mathbf{x}$ as the input, e.g. long short term memory (LSTM)~\cite{hochreiter1997long}.
With the conditional independent assumption (CIA), $P(\pi|\mathbf{x})$ can
be decomposed into a product of posterior $P(\pi_t|\mathbf{x})$ in each frame $t$.

To improve CIA in CTC, RNN transducer~\cite{graves2012sequence}, recurrent neural aligner (RNA)~\cite{sak2017recurrent} and neural segmental model~\cite{tang2017end} are proposed. 
e.g. in RNA, when predicting the  label of the current time step, the predicted label at last time step is used as an additional input to the
recurrent model. An approximate dynamic programming method to optimize the negative
log likelihood, and a sampling-based sequence discriminative
training technique are designed for this model and achieve competitive performance.
With these more powerful inference structures, the label context dependency  can be better modeled and the model works well  without external LM.

Another trend is to use larger model granularities to simplify the training framework and decoding complexity.  
A word level deep learning based acoustic model, A2W CTC~\cite{soltau2016neural}, is trained on 125K hours labeled data and outperforms models with smaller granularities. 
To reduce the amount of acoustic data required to train this model, 
\cite{audhkhasi2017direct} proposes: i) initializing lower layers by mono-phone CTC~\cite{audhkhasi2017direct} to improve  acoustics extraction.
ii) initializing upper  dense layers by GloVe~\cite{pennington2014glove} word embeddings to capture word
co-occurrence information.

\vspace{-1em}  
\subsection{Attention-based Encoder-decoder}
\label{sec:enc-dec-review}
\vspace{-0.5em}  
Attention based encoder-decoder (S2S)~\cite{collobert2016wav2letter,chan2016end,variani2017end,prabhavalkar2017comparison} is another branch of end-to-end model. It does not take CIA in CTC and predicts the posterior probability of the label sequence given both the feature sequence $\mathbf{x}$ and the previous inference labels $\mathbf{l}_{1:i-1}$.
\begin{equation}
\vspace{-1em}  
\label{equ:enc-dec}
\begin{split}
P(\mathbf{l}|\mathbf{x})=\prod_i P(l_{i}|\mathbf{x},\mathbf{l}_{1:i-1})
\end{split}
\end{equation}
An attention mechanism weights the hidden vectors of the feature sequence so that the most related hidden vectors are used for the prediction.
\vspace{-0.5em}  
\begin{equation}
\label{equ:enc-dec-dec}
\vspace{-0.5em}  
\begin{split}
\mathbf{h}&=\text{Encoder}(\mathbf{x})\\
P(l_{i}|\mathbf{x},\mathbf{l}_{1:i-1})&=\text{AttentionDecoder}(\mathbf{h},\mathbf{l}_{1:i-1}) 
\end{split}
\end{equation}
where $\text{Encoder}(\cdot)$ can be LSTM or bidirectional LSTM (BLSTM) and  $\text{AttentionDecoder}(\cdot)$  can be LSTM or gated recurrent unit (GRU).

Recently,  several model structure improvements are proposed.
To help the model convergence, 
a pyramid structure is used in the encoder network to sub-sample the feature sequence~\cite{chan2016end}.
The attention training can be further combined with
CTC criterion in a multi-task learning way, which greatly improves its convergence~\cite{kim2017joint}.
To enable
streaming recognition,  online encoder-decoder structures are proposed, e.g. window based attention~\cite{prabhavalkar2017analysis}, Gaussian prediction based attention~\cite{hou2017gaussian} and reinforcement learning based method~\cite{luo2017learning}.

As speech recognition is both an acoustic and linguistic pattern recognition problem, several works try to utilize text data in the
initialization stage before E2E joint optimization.
The decoder part of S2S can be initialized by a LM trained from text data~\cite{hori2017advances,sriram2017cold}, which helps the model convergence and domain adaptation.

\vspace{-1em}  
\subsection{Drawbacks of Current Frameworks}
\label{sec:drawback}
\vspace{-0.5em}  
Firstly, the acoustic and text data are not utilized  systematically, which results in a large acoustic data requirement.
Current trials merely use text data for initialization, but not utilize both text and acoustic data in a systematical framework. Nevertheless, the traditional neural network based hidden markov model (NN-HMM) framework takes Bayesian theorem and combines acoustic model, transition model, lexicon and LM in a modular way.

Secondly, 
both AM and LM utilize the same model unit, which hurts the generalization and performance. 
In human speech perception, phoneme perception is the foundation of word recognition~\cite{pisoni1985speech}. 
Grapheme and  word  are not designed for acoustics and do not have explicit relationship with their pronunciation. Such trials ignore the prior phoneme knowledge in human language. 
Meanwhile, taking character or grapheme as the model unit for LMs  also deteriorates the performance~\cite{jozefowicz2016exploring}.

Overall speaking, 
E2E systems form a different modeling framework compared with NN-HMM hybrid systems. Hence, prior arts and different knowledge sources of both AM and LM are hard to  be applied in E2E ASR.

\vspace{-0.5em}  
\section{modular-trained neural A2W model}
\label{sec:modu-ete}
\vspace{-0.5em}

\subsection{Training and Decoding Framework}
\label{sec:framework}
\vspace{-0.5em}  

\begin{figure}
  \centering
    \includegraphics[width=\linewidth]{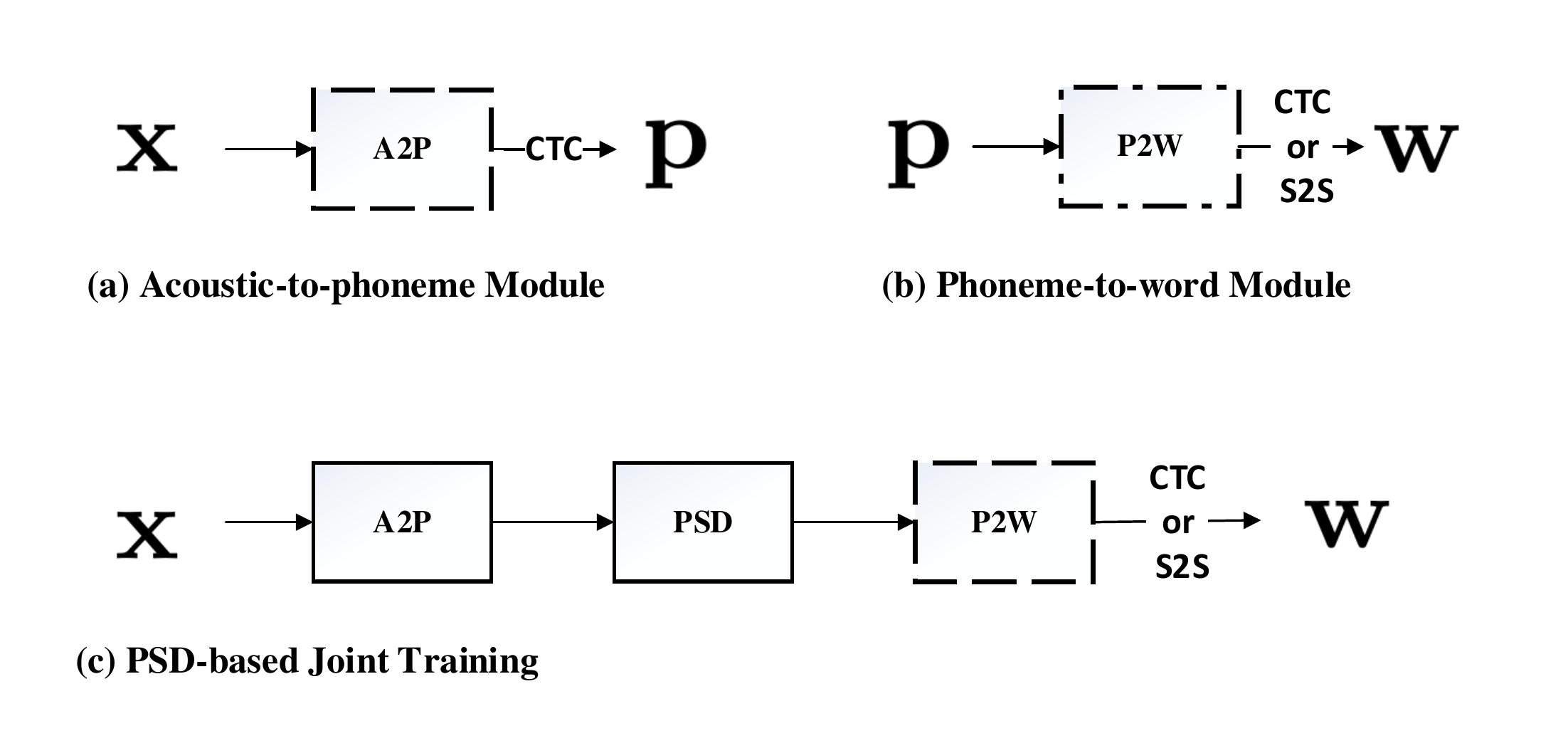}
    \vspace{-2.5em}
    \caption{\it Framework of Modular Training of Neural Acoustics-to-Word Model. The solid line box denotes the layers whose parameters are fixed. The dash line and dash-dot line boxes  denote that  models are trained from acoustic data and text data respectively. }
    \vspace{-1.5em}
    \label{fig:framework}
\end{figure}

Previous works of E2E ASR focus on integrating all components into a whole with both joint optimization and E2E decoding. 
In this work, a modular training strategy is proposed to improve the performance by utilizing external sources to train each building block, while end-to-end decoding is retained to keep the efficient decoding  advantage of E2E ASR. 
Figure~\ref{fig:framework} shows the framework of the proposed method.

The E2E word sequence recognition is modularized as below.
\begin{equation}
\label{equ:framework-1}
P(\mathbf{w}|\mathbf{x})\approx\max_{\mathbf{p}} \left[\ P(\mathbf{w}|\mathbf{p}) \cdot P(\mathbf{p}|\mathbf{x})\ \right]
\end{equation}
where $\mathbf{w}$, $\mathbf{p}$ and $\mathbf{x}$ are word sequence, phoneme sequence and acoustic feature sequence respectively.
An acoustics-to-phoneme model (A2P)   is trained by CTC criterion using acoustic data. 
Meanwhile, a phoneme-to-word model (P2W)   is trained by CTC or S2S using text. 

Then modules are integrated into an A2W model (A2W)  by phone synchronous decoding (PSD)~\cite{zhc00-chen-tasl2017} and joint optimization. 
\begin{equation}
\label{equ:framework-2}
P(\mathbf{w}|\mathbf{x})\approx \max_{\mathbf{p}}\left[\ P(\mathbf{w}|\mathbf{p}) \cdot PSD(\ P(\mathbf{p}|\mathbf{x})\ )\ \right]
\end{equation}

In decoding stage, the jointly optimized A2W model is used as a whole to infer word sequences directly, whose complexity is similar to that of  traditional A2W systems~\cite{audhkhasi2017direct}.
For CTC, the maximum inference labels in each output step are concatenated as the decoding result. Viterbi beam search algorithm is applied in S2S. 
The A2W models can be further combined with external language models to improve the performance. 
In this case, the n-gram language model is compiled to a word WFST.
Thus, PSD search algorithm~\cite{zhc00-chen-tasl2017} can be applied in word level to speed up the pipeline.

\vspace{-1em}  
\subsection{Modularization}
\label{sec:modu}
\vspace{-0.5em}  
As phoneme is a prior knowledge to define all possible  pronunciation the ASR system cares about, 
it  has explicit relationship with acoustics and it is taken as the acoustic model unit.
The A2P module utilizes acoustic data and predicts $P(\mathbf{p}|\mathbf{x})$ as Figure~\ref{fig:framework}(a), which is the same to typical phoneme CTC models~\cite{miao2015eesen}. Notably, although CTC is applied in this work for better comparison purpose, 
other traditional acoustic models, inferring phoneme given acoustics, can also be applied in this module, e.g. RNA~\cite{sak2017recurrent} and LF-MMI~\cite{povey2016purely}.

Different from the end-to-end models discussed in Section~\ref{sec:e2e-review}, the LM here takes the word as the inference unit and predicts $P(\mathbf{w}|\mathbf{p})$ given the phoneme sequence, i.e. phoneme-to-word model in Figure~\ref{fig:framework}(b).
Another key difference is that P2W module utilizes text data and a lexicon, nor acoustic data.
Overall speaking, the P2W module is similar to traditional LMs except that: 
i) P2W consumes phoneme sequences and implicitly  does the tokenization on them.
ii) P2W infers the word sequence given the phoneme sequence. Hence different from traditional LMs, inferring the next word given the previous word sequence, P2W  gets more hints from the next word. Our experiment results also show that P2W performs better than LMs in predicting word sequences.
iii) P2W is trained by sequence criteria, CTC and S2S, which automatically learn the alignment between phoneme sequences and word sequences.

An extra word boundary unit $\tt wb$ is introduced into the phoneme set to improve the tokenization discussed above. $\tt wb$ exists in the end of the phoneme sequence belonging to each word. e.g., the lexicon ``okay ow k ey'' is changed to ``okay ow k ey $\tt wb$''. The motivation is to take $\tt wb$ as the hint of tokenization, e.g. distinguishing short words in case that its phoneme sequence is a substring of longer words.

The advantages of the modularization include: i) both AM and LM use proper inference unit. ii) 
systematically combining acoustic and text data in the model training.

\vspace{-1em}  
\subsection{Phone Synchronous Decoding}
\label{sec:psd}
\vspace{-0.5em}  
With different inference units, namely phoneme and word, PSD is applied to cope with  different information rates.

PSD is originally proposed to speed up the ASR decoding~\cite{zhc00-chen-tasl2017}.
For frames where $\tt blank$
symbol dominates in the CTC inference sequence, it is redundant to do searching as
no phoneme information is provided. Given this observation, phone
synchronous decoding is proposed by skipping the
search of blank-dominated time steps during CTC decoding.
Statistic in~\cite{zhc00-chen-tasl2017} shows that the remaining inference distributions are very compact, which greatly reduces the information rate without precision loss.

Here, $PSD(\cdot)$ is applied as a sub-sampling layer upon the A2P inference sequence, which reduces the input sequence length of P2W layers in the joint optimization. The advantages of PSD module include: i)  ease the  burden of LSTM temporal modeling. ii)  speed up the joint optimization.
Besides, PSD module also speeds up the decoding stage~\cite{zhc00-chen-tasl2017}.

\vspace{-1em}  
\subsection{Joint Optimization}
\label{sec:joint}
\vspace{-0.5em}  
Finally, modules are stacked as Figure~\ref{fig:framework}(c). 
The acoustic data is used to fine-tune the stacked model.
Word level CTC criterion is applied similar to~\cite{soltau2016neural}.
Meanwhile, S2S is  applied in word level for the first time.
During the optimization, only P2W module is fine-tuned and the reason includes: i) the A2P module, mono-phone level CTC model, can always achieve good modeling effects~\cite{miao2015eesen,sak2015fast}. ii) fixing A2P and combining PSD module can greatly speed up the joint optimization. The advantage of sequence modeling in end-to-end model is retained in this stage.

\vspace{-0.8em}  
\section{Relation to Prior Work}
\label{sec:relatework}
\vspace{-0.7em}

End-to-end ASR models always utilize large amount of acoustic data to  optimize the sequence level model jointly. 
e.g., in~\cite{soltau2016neural}, a word level CTC acoustic model is trained on 125K hours labeled data and outperforms hybrid models with smaller granularity. 
As  speech recognition is both an acoustic and linguistic pattern recognition problem, several works try to utilize text data in the initialization stage before end-to-end joint optimization. 
In CTC, \cite{audhkhasi2017direct} proposes to initialize the word output dense
layers with GloVe~\cite{pennington2014glove} word embeddings to capture word co-occurrence information.
In attention-based encoder-decoder, 
the decoder part can be initialized by a LM trained from text data~\cite{hori2017advances,sriram2017cold}, which helps the model convergence and domain adaptation.
This work can be applied in both CTC and S2S. The key differences include: i) modularizing the end-to-end speech recognition by Bayesian theorem. ii) utilizing respective inference units for  acoustic and language modeling. iii) the LM generalizes  word sequences and  lexicons jointly.
Recently, novel structures~\cite{graves2012sequence,sak2017recurrent,tang2017end} are proposed to improve the CIA in CTC by jointly training a LM with the vanilla CTC. A key difference is that the proposed AM and LM infer phonemes and words respectively, which utilizes the linguistic knowledge as discussed in~\ref{sec:modu}.

\vspace{-0.8em}  
\section{Experiments}
\label{sec:exp}
\vspace{-0.8em}


\subsection{Experimental Setup}
\label{sec:exp-setup}
\vspace{-0.5em}  
Experiments were conducted on Switchboard corpus~\cite{godfrey1992switchboard}, which contains about 300 hours of speech. 
36-dimensional filterbank over 25 ms frames every 10 ms from the input speech signal was extracted. Neural networks were trained by Torch~\cite{paszke2017pytorch} and Kaldi~\cite{povey2011kaldi,miao2015eesen}.
The model unit of  phoneme CTC was 45 mono-phones and a $\tt blank$. The baseline phoneme CTC was with 5-layer LSTMs, each with 1024 memory cells and 256 nodes projection layer~\cite{sak2014long}. 
The baseline hybrid system was trained by cross entropy (CE) criterion  and  with the same structure except the last layer, which was with 8K clustered tri-phone states.
The CTC model was initialized by the baseline hybrid system above~\cite{sak2015fast} and the training procedure was similar to \cite{miao2015eesen}. 
In modular training, the baseline phoneme CTC was taken as the A2P module. 
The CTC P2W module was with 4-layer LSTMs, each with 700 memory cells and 256 nodes projection layer. The S2S P2W module used 5-layer LSTMs with 700 nodes for encoder and 1-layer LSTM with 700 nodes for decoder. The vocabulary size was 30K as the standard evaluation setup in this corpus and it was taken as the output layer of P2W module~\footnote{Specific filtering on the vocabulary may improve the performance~\cite{soltau2016neural}. Nevertheless, we didn't do that to make it comparable with other traditional systems in this corpus and also simplify the pipeline.}.
%
%
As a baseline of the end-to-end system without modular training, the direct A2W CTC, similar to~\cite{audhkhasi2017direct}, was with the same structure as the phoneme CTC except the last layer with 30K words. It was initialized by the phoneme CTC.

Evaluation was carried out on the Switchboard (swbd) and Callhome (callhm) subset of the NIST 2000 CTS test set. 
A 30k-vocabulary tri-gram LM trained from the transcription of  Switchboard corpus, without Fisher corpus interpolation, was used for the decoding of baseline phoneme CTC. The P2W module was also trained from the Switchboard transcription to make them comparable with each other.
The decoding procedures of CD-phone CE and CI-phone CTC were the same to~\cite{miao2015eesen}.
The decoding algorithm of A2W systems  was discussed in Section~\ref{sec:framework}. S2S utilized a beam of 20. 
Word error rate (WER) and phoneme error rate (PER) were taken as the metric.

\vspace{-1em}  
\subsection{Modularization}
\label{sec:exp-modu}
\vspace{-0.5em}  
Table~\ref{tab:exp-module} shows the performance of each module in the validation set (CV). Systems with bold-font are used in the later experiments.

\begin{table}[thbp!]
\vspace{-1.7em}  
  \caption{\label{tab:exp-module} {\it  Performance of Each Module. }}
  \centerline{
    \begin{tabular}{c c c c||m{0.15\columnwidth}}
      \hline
      Module  & Model & Inf. Label & Word bound. & PER/WER CV (\%) \\
      \hline \hline
      \multirow{2}{0.08\columnwidth}{A2P}& \multirow{2}{0.08\columnwidth}{CTC} & \multirow{2}{0.15\columnwidth}{phoneme} & $\times$& 13.0 \\
      &  &  & $\surd$& {\bf 12.0} \\
      \hline\hline
      \multirow{4}{0.08\columnwidth}{P2W}& \multirow{2}{0.08\columnwidth}{CTC}  &\multirow{2}{0.1\columnwidth}{word}   & $\times$  &16.0 \\
      &  &  & $\surd$  &{\bf 4.3} \\
      \cline{2-5}
      & \multirow{2}{0.08\columnwidth}{S2S}  & \multirow{2}{0.1\columnwidth}{word}  & $\times$  & 13.9 \\
      &  &  & $\surd$  & {\bf 2.8} \\
      \hline
    \end{tabular}
  }
  \vspace{-1em}  
\end{table}

In A2P, the baseline phoneme recognition performance is parallel with~\cite{graves2006connectionist}. 
$\tt wb$ discussed in Section~\ref{sec:modu} does not hurt the performance. 
The slight improvement stems from the statistic of PER includes $\tt wb$. Further statistic shows  $\tt wb$ prediction error rate is 4\%.

In the P2W module, both CTC and S2S are examined. Without $\tt wb$ inserted in the phoneme sequence, both CTC and S2S obtain  large WERs. 
As discussed in Section~\ref{sec:modu}, $\tt wb$ gives hints of phoneme sequence tokenization. Hence both CTC and S2S with $\tt wb$ improves significantly. 
S2S consistently shows better performance  versus CTC, which benefits from removal of the CIA in CTC~\cite{chan2016end}.
Different from traditional LMs, the  perplexity (PPL) is not reported because the phoneme sequence and the word sequence have different lengths, while the alignment between  them is uncertain and it is automatically learned by sequence criteria.


\vspace{-1em}  
\subsection{Joint Optimization}
\label{sec:exp-joint}
\vspace{-0.5em}  
After modular initialization, models are jointly optimized in Table~\ref{tab:exp-joint}. 
To better support the results, we compare them with the  recent A2W system in this corpus~\cite{audhkhasi2017direct}. 
The different setups include:  
i)  i-vector based adaptation ii)  utilizing BLSTM iii) the LM is interpolated with Fisher corpus. 
Hence the gaps between baselines of this work and those of~\cite{audhkhasi2017direct} are always relative 20-30\%. 

\begin{table}[thbp!]
\vspace{-1.5em}  
  \caption{\label{tab:exp-joint} {\it  Performance Comparison  with or w/o Modular Training. }}
  \centerline{
    \begin{tabular}{c |c |c m{0.15\columnwidth}||c c}
      \hline
       & E2E& \multicolumn{2}{c||}{Modularization } & \multicolumn{2}{c}{WER (\%)}\\
       Name &  Opt. & A2P&P2W & swbd & callhm \\
      \hline \hline
      CD-phone CE & $\times$ & HMM & WFST  &  14.9 & 27.6 \\
      CI-phone CTC & $\times$ & CTC & WFST  & 19.4 & 33.5 \\
      \hline\hline
      \multirow{1}{0.17\columnwidth}{Word CTC}&$\surd$ & n/a & n/a  & 29.6 & 41.7 \\         
      \hline\hline
      \multirow{2}{0.17\columnwidth}{Mod. CTC}&$\surd$ & CTC & CTC  & 24.9 & 36.5 \\   
      &$\surd$ & CTC & \ \ \ \ +WFST  & 23.0 & 35.1 \\ 
      \hline
       \multirow{1}{0.17\columnwidth}{Mod. S2S}&$\surd$ & CTC & S2S  & 31.2 & 40.5 \\   
      \hline
    \end{tabular}
  }
  \vspace{-1em}  
\end{table}

The baseline hybrid system (CD-phone CE) and phoneme CTC (CI-phone CTC) are in line 1 and 2 respectively. Both of them are decoded with a WFST generated from the 30K lexicon and  n-gram LM.
The performance of CI-phone CTC is worse than that of CD-phone CE~\footnote{CTC  performs better than hybrid systems with more data~\cite{sak2015fast}. Our previous work \cite{zhc00-chen-tasl2017} also shares similar finding in a large dataset.}, whose gap is similar to that in~\cite{audhkhasi2017direct}~\footnote{In the same corpus, the phoneme CTC baseline WER in~\cite{audhkhasi2017direct} is 14.5\% and 25.1\% for $\tt swbd$ and $\tt callhm$ respectively. }.
%
The direct A2W CTC (Word CTC) is in line 3. It is with phoneme initialization but without GloVe initialization~\cite{audhkhasi2017direct}. The performance is significantly worse than  CI-phone CTC~\footnote{
The WER  in \cite{audhkhasi2017direct} is 24.4\% and 34.1\%, which shows similar trend.
}.  This setup is taken as the naive  A2W baseline, where Glove  is not implemented as we believe the modular training is a better way to capture the linguistic information.  

The proposed modular-trained A2W CTC (Mod. CTC) is in line 4.
The PSD based joint optimization is applied here, whose effects will be examined  in Table~\ref{tab:exp-psd} later.
Mod. CTC  outperforms the naive A2W in line 3 significantly. 
The modular training framework benefits from: i) easier and faster model convergence due to modularization and initialization. ii) easy to utilize
standard AM and LM technologies using text and acoustic data respectively. 

Table~\ref{tab:exp-psd} shows modeling effects and training speeds with or without PSD. 
All  results are reported on a single Titan GPU. ``fr./s.'' denotes the number of acoustics frames processed per second.   
The training speedup  stems from two folds: i) PSD reduces the sequence length to be processed by P2W in each sequence. ii) As the sequence length is reduced, more sequences can be loaded into GPU memory for parallel training.
Meanwhile, the performance is significantly improved. We believe it also results from  sequence length reduction. Although LSTM is used, the model is still hard to remember a very long input sequence. Nevertheless, for A2W modeling, the history to be remembered before inferring each word is much longer than that of the traditional CI-phone CTC or hybrid systems. Similar problem is reported in~\cite{chan2016end} and solved by a pyramid model structure. The PSD framework shows another choice to cope with this problem.


\begin{table}[thbp!]
\vspace{-1.6em}  
  \caption{\label{tab:exp-psd} {\it  Performance and Speed Comparison  with or w/o PSD. }}
  \centerline{
    \begin{tabular}{c c||c c| c c}
      \hline
        && \multicolumn{2}{c|}{Training Speed} &  \multicolumn{2}{c}{WER (\%)} \\
       Name & PSD & Seq./GPU&  fr./s. & swbd & callhm \\
      \hline \hline
     \multirow{2}{0.2\columnwidth}{Mod. CTC} & $\times$ &5 & 1027 & 32.0 & 42.5 \\
        & $\surd$ &{\bf 30} & {\bf 5851} & {\bf 24.9} &{\bf 36.5}  \\
      \hline
    \end{tabular}
  }
  \vspace{-1em}  
\end{table}

To alleviate the deterioration of the word sequence modeling effect   because of the CIA in CTC, two methods are further investigated~\footnote{We notice that several recent advances of CTC  discussed in Section~\ref{sec:ctc-review} also cope with this problem, which can be our future works.}.
Firstly, a WFST generated from the previous n-gram LM is used to decode with the proposed system. The result is shown in Table~\ref{tab:exp-joint} line 5 with moderate improvement. Hence, the performance gap between line 2 and line 5, CI-phone CTC versus the A2W system, is reduced to relative 15\%~\footnote{Our recent trials on more potent P2W structure finally eliminates the gap.}.
Another method is replacing CTC with S2S discussed in Section~\ref{sec:enc-dec-review}.
The proposed modular-trained A2W S2S (Mod. S2S) is in Table~\ref{tab:exp-joint} the last line. Different from the observation in Table~\ref{tab:exp-module}, the S2S based system does not achieve improvement. Analysis in the decoding result shows that S2S is prone to the phoneme recognition errors from the A2P module. After joint optimization, S2S can not restore the errors, which is similar to the observation in~\cite{prabhavalkar2017comparison}.
Besides, grapheme based systems are not included in this work, although they have been investigated in most of the S2S works, whereas the grapheme is not suitable for language modeling as discussed in Section~\ref{sec:drawback}.

%
%
%
%

%

\vspace{-1em}  
\section{Conclusion}
\label{sec:conclu}
\vspace{-1em}        


In this paper, modular training strategy of E2E ASR is proposed, while end-to-end decoding is retained.
%
Experiments on 300 hours Switchboard corpus show significant improvement on the naive word CTC (A2W).
Compared with the recent A2W system~\cite{audhkhasi2017direct} in this corpus, our performance gap between CI-phone CTC and the A2W model is reduced~\footnote{For $\tt swbd$, 19.4\% to 23.0\% in our setup versus 14.5\% to 21.7 in~\cite{audhkhasi2017direct}.
Meanwhile, other performance gaps show similar trends, e.g. CD-phone CE versus CI-phone CTC, and word CTC versus naive word CTC.}, which also confirms the efficiency of this work.
To further reduce the gap, future works in the modular training framework include: i) utilizing external knowledge sources. ii) applying novel models~\cite{graves2012sequence,sak2017recurrent}. iii) improving  S2S-based P2W.

\vfill\pagebreak

\bibliographystyle{IEEEbib}
\bibliography{refs}

\begin{thebibliography}{10}

\bibitem{sak2014long}
Ha{\c{s}}im Sak, Andrew Senior, and Fran{\c{c}}oise Beaufays,
\newblock ``Long short-term memory recurrent neural network architectures for
  large scale acoustic modeling,''
\newblock in {\em Fifteenth Annual Conference of the International Speech
  Communication Association}, 2014.

\bibitem{qian2016very}
Yanmin Qian, Mengxiao Bi, Tian Tan, and Kai Yu,
\newblock ``Very deep convolutional neural networks for noise robust speech
  recognition,''
\newblock {\em IEEE/ACM Transactions on Audio, Speech, and Language
  Processing}, vol. 24, no. 12, pp. 2263--2276, 2016.

\bibitem{sak2015fast}
Ha{\c{s}}im Sak, Andrew Senior, Kanishka Rao, and Fran{\c{c}}oise Beaufays,
\newblock ``Fast and accurate recurrent neural network acoustic models for
  speech recognition,''
\newblock {\em arXiv preprint arXiv:1507.06947}, 2015.

\bibitem{soltau2016neural}
Hagen Soltau, Hank Liao, and Hasim Sak,
\newblock ``Neural speech recognizer: Acoustic-to-word lstm model for large
  vocabulary speech recognition,''
\newblock {\em arXiv preprint arXiv:1610.09975}, 2016.

\bibitem{graves2012sequence}
Alex Graves,
\newblock ``Sequence transduction with recurrent neural networks,''
\newblock {\em arXiv preprint arXiv:1211.3711}, 2012.

\bibitem{sak2017recurrent}
Ha{\c{s}}im Sak, Matt Shannon, Kanishka Rao, and Fran{\c{c}}oise Beaufays,
\newblock ``Recurrent neural aligner: An encoder-decoder neural network model
  for sequence to sequence mapping,''
\newblock {\em Proc. Interspeech 2017}, pp. 1298--1302, 2017.

\bibitem{tang2017end}
Hao Tang, Liang Lu, Lingpeng Kong, Kevin Gimpel, Karen Livescu, Chris Dyer,
  Noah~A Smith, and Steve Renals,
\newblock ``End-to-end neural segmental models for speech recognition,''
\newblock {\em IEEE Journal of Selected Topics in Signal Processing}, 2017.

\bibitem{collobert2016wav2letter}
Ronan Collobert, Christian Puhrsch, and Gabriel Synnaeve,
\newblock ``Wav2letter: an end-to-end convnet-based speech recognition
  system,''
\newblock {\em arXiv preprint arXiv:1609.03193}, 2016.

\bibitem{chan2016end}
William Chan,
\newblock {\em End-to-End Speech Recognition Models},
\newblock Ph.D. thesis, Carnegie Mellon University Pittsburgh, PA, 2016.

\bibitem{zhc00-chen-tasl2017}
Z.~Chen, Y.~Zhuang, Y.~Qian, and K.~Yu,
\newblock ``Phone synchronous speech recognition with ctc lattices,''
\newblock {\em IEEE/ACM Transactions on Audio, Speech, and Language
  Processing}, vol. 25, no. 1, pp. 86--97, Jan 2017.

\bibitem{graves2006connectionist}
Alex Graves, Santiago Fern{\'a}ndez, Faustino Gomez, and J{\"u}rgen
  Schmidhuber,
\newblock ``Connectionist temporal classification: labelling unsegmented
  sequence data with recurrent neural networks,''
\newblock in {\em Proceedings of the 23rd international conference on Machine
  learning}. ACM, 2006, pp. 369--376.

\bibitem{hochreiter1997long}
Sepp Hochreiter and J{\"u}rgen Schmidhuber,
\newblock ``Long short-term memory,''
\newblock {\em Neural computation}, vol. 9, no. 8, pp. 1735--1780, 1997.

\bibitem{audhkhasi2017direct}
Kartik Audhkhasi, Bhuvana Ramabhadran, George Saon, Michael Picheny, and David
  Nahamoo,
\newblock ``Direct acoustics-to-word models for english conversational speech
  recognition,''
\newblock {\em arXiv preprint arXiv:1703.07754}, 2017.

\bibitem{pennington2014glove}
Jeffrey Pennington, Richard Socher, and Christopher Manning,
\newblock ``Glove: Global vectors for word representation,''
\newblock in {\em Proceedings of the 2014 conference on empirical methods in
  natural language processing (EMNLP)}, 2014, pp. 1532--1543.

\bibitem{variani2017end}
Ehsan Variani, Tom Bagby, Erik McDermott, and Michiel Bacchiani,
\newblock ``End-to-end training of acoustic models for large vocabulary
  continuous speech recognition with tensorflow,''
\newblock {\em Proc. Interspeech 2017}, pp. 1641--1645, 2017.

\bibitem{prabhavalkar2017comparison}
Rohit Prabhavalkar, Kanishka Rao, Tara~N Sainath, Bo~Li, Leif Johnson, and
  Navdeep Jaitly,
\newblock ``A comparison of sequence-to-sequence models for speech
  recognition,''
\newblock {\em Proc. Interspeech 2017}, pp. 939--943, 2017.

\bibitem{kim2017joint}
Suyoun Kim, Takaaki Hori, and Shinji Watanabe,
\newblock ``Joint ctc-attention based end-to-end speech recognition using
  multi-task learning,''
\newblock in {\em Acoustics, Speech and Signal Processing (ICASSP), 2017 IEEE
  International Conference on}. IEEE, 2017, pp. 4835--4839.

\bibitem{prabhavalkar2017analysis}
Rohit Prabhavalkar, Tara~N Sainath, Bo~Li, Kanishka Rao, and Navdeep Jaitly,
\newblock ``An analysis of “attention” in sequence-to-sequence models,''
\newblock {\em Proc. Interspeech 2017}, pp. 3702--3706, 2017.

\bibitem{hou2017gaussian}
Junfeng Hou, Shiliang Zhang, and Lirong Dai,
\newblock ``Gaussian prediction based attention for online end-to-end speech
  recognition,''
\newblock {\em Proc. Interspeech 2017}, pp. 3692--3696, 2017.

\bibitem{luo2017learning}
Yuping Luo, Chung-Cheng Chiu, Navdeep Jaitly, and Ilya Sutskever,
\newblock ``Learning online alignments with continuous rewards policy
  gradient,''
\newblock in {\em Acoustics, Speech and Signal Processing (ICASSP), 2017 IEEE
  International Conference on}. IEEE, 2017, pp. 2801--2805.

\bibitem{hori2017advances}
Takaaki Hori, Shinji Watanabe, Yu~Zhang, and William Chan,
\newblock ``Advances in joint ctc-attention based end-to-end speech recognition
  with a deep cnn encoder and rnn-lm,''
\newblock {\em arXiv preprint arXiv:1706.02737}, 2017.

\bibitem{sriram2017cold}
Anuroop Sriram, Heewoo Jun, Sanjeev Satheesh, and Adam Coates,
\newblock ``Cold fusion: Training seq2seq models together with language
  models,''
\newblock {\em arXiv preprint arXiv:1708.06426}, 2017.

\bibitem{pisoni1985speech}
David~B Pisoni, Howard~C Nusbaum, Paul~A Luce, and Louisa~M Slowiaczek,
\newblock ``Speech perception, word recognition and the structure of the
  lexicon,''
\newblock {\em Speech communication}, vol. 4, no. 1-3, pp. 75--95, 1985.

\bibitem{jozefowicz2016exploring}
Rafal Jozefowicz, Oriol Vinyals, Mike Schuster, Noam Shazeer, and Yonghui Wu,
\newblock ``Exploring the limits of language modeling,''
\newblock {\em arXiv preprint arXiv:1602.02410}, 2016.

\bibitem{miao2015eesen}
Yajie Miao, Mohammad Gowayyed, and Florian Metze,
\newblock ``Eesen: End-to-end speech recognition using deep rnn models and
  wfst-based decoding,''
\newblock in {\em Automatic Speech Recognition and Understanding (ASRU), 2015
  IEEE Workshop on}. IEEE, 2015, pp. 167--174.

\bibitem{povey2016purely}
Daniel Povey, Vijayaditya Peddinti, Daniel Galvez, Pegah Ghahrmani, Vimal
  Manohar, Xingyu Na, Yiming Wang, and Sanjeev Khudanpur,
\newblock ``Purely sequence-trained neural networks for asr based on
  lattice-free mmi,''
\newblock {\em Submitted to Interspeech}, 2016.

\bibitem{godfrey1992switchboard}
John~J Godfrey, Edward~C Holliman, and Jane McDaniel,
\newblock ``Switchboard: Telephone speech corpus for research and
  development,''
\newblock in {\em Acoustics, Speech, and Signal Processing, 1992. ICASSP-92.,
  1992 IEEE International Conference on}. IEEE, 1992, vol.~1, pp. 517--520.

\bibitem{paszke2017pytorch}
Adam Paszke, Sam Gross, and Soumith Chintala,
\newblock ``Pytorch,'' 2017.

\bibitem{povey2011kaldi}
Daniel Povey, Arnab Ghoshal, Gilles Boulianne, Lukas Burget, Ondrej Glembek,
  Nagendra Goel, Mirko Hannemann, Petr Motlicek, Yanmin Qian, Petr Schwarz,
  et~al.,
\newblock ``The kaldi speech recognition toolkit,''
\newblock in {\em IEEE 2011 workshop on automatic speech recognition and
  understanding}. IEEE Signal Processing Society, 2011, number
  EPFL-CONF-192584.

\end{thebibliography}

\end{document}